\documentclass{article}

\usepackage[final]{nips_2019}
\usepackage[T1]{fontenc}    
\usepackage{hyperref}       
\usepackage{url}            
\usepackage{subcaption} 
\usepackage{graphicx}
\usepackage{booktabs}       
\usepackage{amsfonts}       
\usepackage{nicefrac}       
\usepackage{microtype}      
\usepackage{natbib}
\usepackage[utf8]{inputenc}
\title{FlowDB: A new large scale river flow, flash flood, and precipitation dataset}
\author{Isaac Godfried, Kriti Mahajan, Maggie Wang, Kevin Li, Pranjalya Tiwari \\ CoronaWhy}
\date{October 2020}

\begin{document}

\maketitle
\begin{abstract}
  Flooding results in 8 billion dollars of damage annually in the US and causes the most deaths of any weather related event. Due to climate change scientists expect more heavy precipitation events in the future. However, no current datasets exist that contain both hourly precipitation and river flow data. We introduce a novel hourly river flow and precipitation dataset and a second subset of flash flood events with damage estimates and injury counts. Using these datasets we create two challenges (1) general stream flow forecasting and (2) flash flood damage estimation. We have created several publicly available benchmarks and an easy to use package. Additionally, in the future we aim to augment our dataset with snow pack data and soil index moisture data to improve predictions. 
\end{abstract}
\section{Introduction}
Flooding has significant negative effects on infrastructure, agriculture, housing and the population as a whole. Approximately 41 million residents in the United States alone are at risk of stream and river flooding. \footnote{https://www.nrdc.org/stories/flooding-and-climate-change-everything-you-need-know} Flooding related costs in the U.S. averaged 8 billion dollars per year between 2011-2018. Additionally, worldwide between 1980 and 2009 there were over 500,000 flood related deaths. While directly linking increased flooding to climate change is difficult, recent research has indicated a statistically significant increase in heavy precipitation events \footnote{\url{https://www.ipcc.ch/site/assets/uploads/2018/03/SREX-Chap3_FINAL-1.pdf} p.111}. Moreover, researchers believe that this trend could continue well into the 21st century \footnote{ibid p.112}. Therefore, it is important to develop accurate forecasts and response plans to help mitigate the damage. 

Although a number of studies have used machine learning to forecast river flows they are usually focused on a single river basin \citet{YASEEN2019387} or a single state \citet{IowaFlood}.  A few studies have applied machine learning to predicting tornado damage \cite{article}. But, we were unable to find any previous research that forecasted flood based property damage. The dataset by \cite{camels} remains a commonly used benchmark, however it only contains roughly 600 streams and data is only provided on the daily basis. Many flash floods develop quite quickly in response to heavy precipitation events therefore a more granular dataset is needed to truly provide timely alerts.

To rectify these problems we purpose a new large-scale benchmark dataset consisting of hourly USGS flow data from rivers around the country combined with hourly precipitation and temperature data. Additionally, we utilize NLP to construct a second subset consisting of flash flood events. This dataset includes flow data, precipitation data, property damage estimates, injury data, and regional information.

\section{Dataset Creation and Prediction Tasks}
 To create a dataset we utilize publicly available stream flow data from the United States Geological Service (USGS). We built a scraper to initially get data over a five year period from 2014-2019 for each USGS gage. For precipitation data we acquired data from USGS, SNOTEL, NOAA, ASOS, EcoNet, as well as other sources. 

In order to get the data together we normalized timestamps and then left-joined the stream flow and the weather data together. We created code to compute the distance between every river gage and weather station. For missing values we use simple mean of the two closest known values on each side. We choose this method as rainfall amount is often directly related to its nearby values. We add a column which indicates whether the value was a result of imputation. Another source of noise in the dataset is the presence of dam fed/controlled rivers. While some dam controlled rivers are easy to identify (due to their cyclical release nature) others are much harder. However, for the time being we elect to keep all river/stream data in the dataset as it could be beneficial to see if the model could accurately forecast release levels. 

To construct the flash flood dataset for the second task, we use NOAA's NSSL Flash dataset \footnote{https://blog.nssl.noaa.gov/flash/database/database-2016v1/}. Similar to the flood forecast we tie the data to rainfall event using timestamps. We use Flair NER \footnote{https://github.com/zalandoresearch/flair} to extract stream names from the narrative reports and when available attach the relevant USGS data.

\section{Dataset statistics}
Task one includes combined river flows, precipitation, and temperature data for more than 8000 gages, in all 50 U.S. states. In total this is over 70 million hours of river flow data. For the second dataset we have a total of 4320 flash flood related events. There is a class imbalance in this dataset particularly with respect to injuries/fatalities with 4262 of the flash flood events resulting in none. With respect to property/crop damage, 2763 of the flash flood events resulted in no damage. While, only 156 events caused over a million dollars in damage. For more detailed information on the flash flood dataset we have an exploratory analysis notebook \footnote{https://rb.gy/wlcg2j}.

\section{Evaluation, Platform, and Benchmark Results}
For task one we choose mean squared error $\frac{1}{n}\sum_{t=1}^{n}(Y_t-Y^{a}_t)^2$ (MSE) as a metric for evaluating individual rivers. Mean squared error is a commonly used metric for time series data and river forecasting \cite{FloodML}. However, since cubic feet per second values vary considerably between rivers simply computing a mean MSE across all rivers unfairly hurts models when they make forecasting errors on large rivers, but under penalizes them for errors on smaller creeks/tributaries. We therefore compute an $R^2$ value for each river. For the $R^2$ value we simply compare the performance of the model versus a baseline $1-\frac{MSE(MODEL)}{MSE(BASELINE)}$. Here the baseline is simply the median stream flow value. Then using this normalized score we can take a simple mean to get the overall score of the model across all rivers. $score_{model}=\frac{1}{i}\sum_{i=1}^{i}R_i^2$ where $i$ is the total number of rivers in the dataset.

We maintain a hold out set of 14 days or 336 hours. We choose 14 days as this is the max length of many weather forecasts. We note that there is a possible train/production drift as this test set is based on the actual rainfall amounts rather than the amounts in the weather forecast. However, for this iteration of the dataset we think that it is beneficial to decouple flow forecasting from rainfall forecasting. This allows us to see how well models are able to learn the actual relationship between rainfall, temperature and other factors absent the noise associated with rainfall forecasts. Future tasks might challenge researchers to perform joint precipitation and flow forecasting.

For our baseline deep learning method we use a dual stage attention based RNN (DA-RNN) \cite{Qin2017ADA}. DA-RNN is a well known model for time-series forecasting that achieved state of the art results on the SML 2010 dataset and the NASDAQ 100 Stock when it was first released. We use the public implementation provided by "Seanny123" \footnote{https://github.com/Seanny123/da-rnn}. Currently, we have only trained/evaluated this model on a several rivers and not the whole dataset. However, for the few rivers we test on we found that even without significant optimization of hyper-parameters that the model often outperformed the median baseline method described above. For instance, for Carter Creek in Sebring FL when forecasting cfs we get a MSE of $0.120$ compared to a baseline MSE of 486.587. This gives us an $R^2$ value .99.  However, for the Kenduskeag stream the model struggles to learn the relevant features for forecasting flow and has an MSE of 22112, however this still beats the median baseline of 23888. 

On task two there are two specific sub-tasks: forecasting property/crop damage and forecasting injuries/fatalities. For injuries we make it into a binary classification problem where the label is 0 if there are no injuries or fatalities and 1 if there are. Hence we use F1 score as our evaluation metric. For property damage due to the high dimensionality of the data and relatively small dataset we bucket damages into categories. $label \in \{0,1,2,3,4\}$ 
\begin{center}
 \begin{tabular}{||c c c c||} 
 \hline
 Label & Min Damage (inclusive) & Max Damage & Count \\ [0.5ex] 
 \hline\hline
 0 & 0 & 0 & 2763 \\ 
 \hline
 1 & 1 & 10,000 & 403 \\
 \hline
 2 & 10,000 & 100,000 & 730 \\
 \hline
 3 & 100,000 & 1,000,000 & 267 \\
 \hline
 4 & 1,000,000 & $\infty$ & 156 \\ [1ex] 
 \hline
\end{tabular}
\end{center}

For evaluation for this task we also use F1 score. We split the dataset randomly into 20\% test and 80\% training.

For the task two challenge we are still researching an appropriate architecture to make the predictions. The architecture will have to incorporate dynamic precipitation data as well as static information such as population and geospatial data on the area.

In order to enable researchers to quickly train/test different configurations with limited overhead we have created a package that provides the basic pre-processing code, our baseline models, and evaluation scripts\footnote{https://pypi.org/project/flood-forecast/}. Our model evaluation script supports the tracking of data changes for scoring purposes. So for instance, when we add additional data such as snow pack and soil it will be easy to tell which version of the dataset was used. Lastly, we structure our package so it is easy to extend models, experiment with different groupings of rivers for transfer learning, and create new models altogether. \footnote{https://github.com/AIStream-Peelout/flow-forecast}

\section{Continued Work}
There are several areas for improvement that we hope to address in future iterations of the dataset. Specifically, we hope to incorporate snow pack depth and soil moisture index to help improve forecasts in mountain areas. Secondly, we are currently expanding our dataset to include data from 1980-2014 so that researchers can see in more detail the affects of climate change. Finally, there are a number of interesting research questions that researchers could use this dataset to investigate. These include investigating transfer, continual, and few shot learning for time series forecasting. For instance, we also found that pre-training time series models on the river flow data improved performance at forecasting COVID-19 spread. \footnote{https://rb.gy/ejwufr}

In summary, our main contributions include: creating the only hourly publicly available dataset that contains both precipitation and river/stream flows (and includes more than ten times the number of rivers/streams as \cite{camels}), making a simple set of benchmarks with DA-RNN, and providing an easy to use framework for researchers to test new models/features on this challenging problem.

\bibliography{sources}
\bibliographystyle{plainnat}
\end{document}